\documentclass[10pt,twocolumn,letterpaper]{article}

\usepackage[pagenumbers]{cvpr} 

\definecolor{cvprblue}{rgb}{0.21,0.49,0.74}
\usepackage[pagebackref,breaklinks,colorlinks,allcolors=cvprblue]{hyperref}
\usepackage{graphicx}
\usepackage{amsmath}
\usepackage{amssymb}
\usepackage{tabularx,booktabs}
\newcolumntype{Y}{>{\centering\arraybackslash}X} 

\usepackage{cuted} 
\usepackage{xcolor}

\usepackage{array}
\usepackage{subcaption}
\usepackage{stfloats}
\usepackage{placeins}

\usepackage[pagebackref,breaklinks,colorlinks]{hyperref}
\usepackage[capitalize]{cleveref}

\usepackage{comment}
\usepackage{amsmath,amssymb} %
\usepackage{wrapfig} %
\usepackage{color}

\usepackage{mathtools}
\usepackage[accsupp]{axessibility}  %

\usepackage[utf8]{inputenc} %
\usepackage[T1]{fontenc}    %
\usepackage{xr-hyper}
\usepackage{url}            %
\usepackage{booktabs}       %
\usepackage{xcolor,colortbl}         %
\usepackage{caption}

\usepackage{multirow}
\usepackage{enumitem}

\usepackage[normalem]{ulem}

\usepackage{subcaption}
\usepackage{float}
\usepackage{lscape}     

\usepackage{xspace}
\usepackage{setspace}
\usepackage{pifont}
\usepackage{comment}

\usepackage{enumitem}

\def\paperID{} 
\def\confName{CVPR}
\def\confYear{2026}

\title{Intelligent Traffic Surveillance for Real-Time Vehicle Detection, License Plate Recognition, and Speed Estimation}

\author{
Bruce Mugizi \quad
Sudi Murindanyi \quad
Olivia Nakacwa \quad
Andrew Katumba \\
Makerere University
}

\begin{document}


\newcommand{\TODO}[1]{\textcolor{red}{TODO: #1}}
\newcommand{\ab}[1]{{\color{orange}#1}}
\newcommand{\AB}[1]{{\color{cyan}(Ananta:#1)}}
\newcommand{\hr}[1]{{\color{blue}#1}}
\newcommand{\HR}[1]{{\color{blue}(Helge:#1)}}
\newcommand{\NEW}[1]{{\color{blue}#1}}


\newcommand{\tbf}[1]{\textbf{#1}}
\newcommand{\topic}[1]{\textbf{#1}}
\newcommand{\figref}[1]{Figure~\ref{#1}}
\newcommand{\secref}[1]{Section~\ref{#1}}
\newcommand{\feqref}[1]{Equation~\eqref{#1}}
\newcommand{\tabref}[1]{Table~\ref{#1}}

\newcommand{\app}{appendix} 

\newcommand{\parag}[1]{\noindent\textbf{#1}}


\newcommand{\R}{\mathbb{R}}

\newcommand{\va}{\mathbf{a}}
\newcommand{\vb}{\mathbf{b}}
\newcommand{\vc}{\mathbf{c}}
\newcommand{\vd}{\mathbf{d}}
\newcommand{\ve}{\mathbf{e}}
\newcommand{\vf}{\mathbf{f}}
\newcommand{\vg}{\mathbf{g}}
\newcommand{\vh}{\mathbf{h}}
\newcommand{\vi}{\mathbf{i}}
\newcommand{\vj}{\mathbf{j}}
\newcommand{\vk}{\mathbf{k}}
\newcommand{\vl}{\mathbf{l}}
\newcommand{\vm}{\mathbf{m}}
\newcommand{\vn}{\mathbf{n}}
\newcommand{\vo}{\mathbf{o}}
\newcommand{\vp}{\mathbf{p}}
\newcommand{\vq}{\mathbf{q}}
\newcommand{\vr}{\mathbf{r}}
\newcommand{\vt}{\mathbf{t}}
\newcommand{\vu}{\mathbf{u}}
\newcommand{\vv}{\mathbf{v}}
\newcommand{\vw}{\mathbf{w}}
\newcommand{\vx}{\mathbf{x}}
\newcommand{\vy}{\mathbf{y}}
\newcommand{\vz}{\mathbf{z}}

\newcommand{\mA}{\mathbf{A}}
\newcommand{\mB}{\mathbf{B}}
\newcommand{\mC}{\mathbf{C}}
\newcommand{\mD}{\mathbf{D}}
\newcommand{\mE}{\mathbf{E}}
\newcommand{\mF}{\mathbf{F}}
\newcommand{\mG}{\mathbf{G}}
\newcommand{\mH}{\mathbf{H}}
\newcommand{\mI}{\mathbf{I}}
\newcommand{\mJ}{\mathbf{J}}
\newcommand{\mK}{\mathbf{K}}
\newcommand{\mL}{\mathbf{L}}
\newcommand{\mM}{\mathbf{M}}
\newcommand{\mN}{\mathbf{N}}
\newcommand{\mO}{\mathbf{O}}
\newcommand{\mP}{\mathbf{P}}
\newcommand{\mQ}{\mathbf{Q}}
\newcommand{\mR}{\mathbf{R}}
\newcommand{\mS}{\mathbf{S}}
\newcommand{\mT}{\mathbf{T}}
\newcommand{\mU}{\mathbf{U}}
\newcommand{\mV}{\mathbf{V}}
\newcommand{\mW}{\mathbf{W}}
\newcommand{\mX}{\mathbf{X}}
\newcommand{\mY}{\mathbf{Y}}
\newcommand{\mZ}{\mathbf{Z}}

\newcommand{\cA}{\mathcal A}
\newcommand{\cB}{\mathcal B}
\newcommand{\cC}{\mathcal C}
\newcommand{\cD}{\mathcal D}
\newcommand{\cE}{\mathcal E}
\newcommand{\cF}{\mathcal F}
\newcommand{\cG}{\mathcal G}
\newcommand{\cH}{\mathcal H}
\newcommand{\cI}{\mathcal I}
\newcommand{\cJ}{\mathcal J}
\newcommand{\cK}{\mathcal K}
\newcommand{\cL}{\mathcal L}
\newcommand{\cM}{\mathcal M}
\newcommand{\cN}{\mathcal N}
\newcommand{\cO}{\mathcal O}
\newcommand{\cP}{\mathcal P}
\newcommand{\cQ}{\mathcal Q}
\newcommand{\cR}{\mathcal R}
\newcommand{\cS}{\mathcal S}
\newcommand{\cT}{\mathcal T}
\newcommand{\cU}{\mathcal U}
\newcommand{\cV}{\mathcal V}
\newcommand{\cW}{\mathcal W}
\newcommand{\cX}{\mathcal X}
\newcommand{\cY}{\mathcal Y}
\newcommand{\cZ}{\mathcal Z}

\newcommand{\bR}{\mathbb{R}}
\newcommand{\mx}{\mathbf{x}}
\newcommand{\mj}{\mathbf{j}}
\newcommand{\mb}{\mathbf{b}}
\newcommand{\vmu}{\mathbf{\mu}}

\newcommand{\mIrradiance}{\mI_i}
\newcommand{\mAlbedo}{\mA}
\newcommand{\mDepth}{\mD}
\newcommand{\mShadow}{\mS}
\newcommand{\mDiffuse}{\vc_d}
\newcommand{\mAmbient}{\vc_a}
\newcommand{\mLight}{\vn_l}
\newcommand{\mRGB}{\mathbf{RGB}}
\newcommand{\mRGBlit}{\mathbf{RGB_{lit}}}
\newcommand{\normal}{\vn_s} 

\newcommand{\dotprod}{\boldsymbol{\cdot}}

\definecolor{Gray}{gray}{0.85}

\definecolor{DeepGreen}{rgb}{0.15,0.60,0.15}
\newcommand{\canon}[1]{#1^c}
\newcommand{\world}[1]{#1^w}
\newcommand{\Ltwo}[1]{\vert\vert #1\vert\vert^2_2}

\newcommand{\obs}[1]{#1^\vo}

\twocolumn[{
\maketitle
\centering
}]
\begin{abstract}
Speeding is a major contributor to road fatalities, particularly in developing countries such as Uganda, where road safety infrastructure is limited. This study proposes a real-time intelligent traffic surveillance system tailored to such regions, using computer vision techniques to address vehicle detection, license plate recognition, and speed estimation. The study collected a rich dataset using a speed gun, a Canon Camera, and a mobile phone to train the models. License plate detection using YOLOv8 achieved a mean average precision (mAP) of 97.9\%. For character recognition of the detected license plate, the CNN model got a character error rate (CER) of 3.85\%, while the transformer model significantly reduced the CER to 1.79\%. Speed estimation used source and target regions of interest, yielding a good performance of ±10 km/h margin of error. Additionally, a database was established to correlate user information with vehicle detection data, enabling automated ticket issuance via SMS via Africa's Talking API. This system addresses critical traffic management needs in resource-constrained environments and shows potential to reduce road accidents through automated traffic enforcement in developing countries where such interventions are urgently needed.
\end{abstract}    
\section{Introduction}
\label{sec:intro}

Every year, a significant number of lives are lost due to traffic law violations, particularly over-speeding. According to the World Health Organization (WHO), road traffic crashes claim approximately 1.19 million lives globally each year \cite{WHO2023, WHOTopicDetails}, with an additional 20 to 50 million people suffering non-fatal injuries, predominantly in low- and middle-income countries \cite{UNDRR2023, WHO2023}. Various factors contribute to the increased risk of road traffic crashes and fatalities, with high-speed driving notably increasing both the likelihood and severity of crashes \cite{Alnawmasi2022, WAGov2024, FHWA, turyahabwa2025integrative}. The risk of pedestrian fatalities rises sharply with speed; for instance, there is a 4.5-fold greater likelihood of death when struck by high-speed vehicles \cite{ugandaroadsleave200}.

Countries in Africa, primarily developing ones, have the highest traffic-related death rates globally despite having the lowest motorization levels \cite{segui-gomez2021road, Bachani2017}. Over recent decades, without effective road safety programs, mortality from road traffic injuries has steadily increased in regions such as East Asia, South Asia, and Sub-Saharan Africa \cite{Bachani2017, heydari2019road}. In these developing countries, police enforce traffic laws through targeted operations, often using speed guns to monitor speeding \cite{humansocialEnochroad, norbury2020roads, kcca2022kampala}. However, road crashes continue to occur frequently, and these countries experience road traffic injury mortality rates more than double those of developed nations \cite{hyder2003inequality, heydari2019road}. In contrast, developed countries have seen a decline in road traffic fatalities following the implementation of safety programs over the past decade \cite{Bachani2017}. This disparity highlights the need for more comprehensive road traffic laws and more vigorous enforcement in non-developed countries. Effective enforcement of speeding regulations is crucial for reducing crashes and fatalities. 

Uganda, a developing country in eastern Africa, ranks third among countries with the highest traffic death rates, estimated at 29 car deaths per 100,000 people, with road traffic accidents among the top 10 causes of death in the country \cite{Walekhwa2022, healthprofileworldlifeexpectancy, uniph2024}. The Uganda Police indicates an increase in traffic accidents on Ugandan roads, with speeding being the leading cause of crashes \cite{kcca2022kampala}, and according to the Uganda Police Force Annual Crime Report, speeding ranks the highest among the causes of road crashes, accounting for 50\% of total crashes registered \cite{Nankinga2023, annualpolice2023}. Uganda has laid out strategies to reduce road crashes, including capacity building, road safety awareness, digitization of the road crash database system, and community engagements \cite{annualpolice2023}. Additionally, there are four fundamental values of road safety, usually termed the 4E's: Enforcement, Emergency, Engineering, and Education. These have become the main thrust of accident prevention worldwide \cite{MoRTH, MHA2020,4e_consultants}. In education, various road safety campaigns use audiovisual and other print media, as well as non-governmental organizations, to raise awareness. Governments usually undertake various publicity measures through television and radio, distributing posters, books on road safety, signage, and signs to raise road safety awareness among the general public \cite{Imokola2024, RSA, rwandapolice2012safer}. The urgency of using the latest technologies to detect traffic violations and penalize violators has increased with enforcement \cite{ec2018speed}. 
In engineering, efficient methods, which are a combination of technology and skills, can help to improve safety measures. In this paper, we tackle this problem using emerging technologies, such as deep learning. The idea is to implement an automated ticketing system based on deep learning to issue traffic speed-limit violation tickets.

We propose a system with four sub-components: vehicle detection, plate detection, speed estimation, and ticket generation. Vehicle detection analyzes frames to identify vehicles, and speed estimation and plate detection algorithms run using tracking data to estimate speeds and extract plate characters through optical character recognition (OCR). The system integrates the components to generate output data for logging, ticketing, and visuals on which the ticket issuance is based. OCR extracts and identifies text from unstructured documents, such as images and screenshots. Given an extracted image of a license plate, an OCR program extracts the text from that image. 


\section{Related Work}
\label{sec2}
Numerous systems have been developed to address high-speed traffic challenges, using various technologies to improve law enforcement and road safety. These systems range from traditional methods to advanced solutions, including computer vision for vehicle identification and real-time monitoring algorithms. For instance, Seamless Speed Enforcement by VITRONIC uses LIDAR technology to detect speed violations across multiple lanes, even in heavy traffic \cite{vitronicSeamlessSpeed}. Another example is Fixed Spot Speed Enforcement by Sensys Gatso Group \cite{sensysgatsoReliableFixed}. This literature review focuses on recent research aimed at enhancing these systems and improving traffic management and speed regulation enforcement.

\subsection{Vehicle Detection and Tracking}
Vehicle detection and tracking are crucial in computer vision and intelligent transportation systems. Various methods have been developed for this purpose. Wang and zhang \cite{wang2022real} explored 3D LiDAR for real-time vehicle detection and tracking, enhancing precision in urban traffic by combining traditional techniques with point-cloud-based methods. Their approach showed good speed and accuracy under challenging conditions, outperforming image-only methods. Iqra Nosheen et al. \cite{nosheen2024efficient} introduced a system that utilizes blob detection and a Kernelized Correlation Filter (KCF) for tracking. They enhanced accuracy in varying lighting conditions through preprocessing techniques, achieving 82\% accuracy in vehicle detection and 86\% in tracking on the KITTI dataset. Tan et al. \cite{Tan_2020_CVPR} presented EfficientDet, a scalable object detection model that reduces computational costs while maintaining high accuracy. It achieved state-of-the-art performance on COCO benchmarks, making it suitable for low-resource environments like edge devices. Zhen He and Hangen He \cite{he2018unsupervised} developed an end-to-end unsupervised Multi-object Detection (MOD) framework for video surveillance. Their model, employing a Memory-Based Recurrent Attention Network (MRAN), minimizes image reconstruction error to detect multiple objects without labelled training data. Applied to synthetic and real datasets like DukeMTMC, their method outperformed traditional models, achieving an AP of 87.2\%. This approach advances unsupervised video surveillance and multi-object tracking by eliminating the need for manually labelled data.

\subsection{License Plate Detection}
License plates are unique to each vehicle and differ by country. Recent advancements in machine learning have significantly enhanced license plate detection methods. Kothai G. et al. \cite{kothai2024efficient} proposed a CNN-based method for detecting and recognizing license plates, which involves detecting plate contours and applying edge detection for character extraction. This approach has shown improved precision in real-time scenarios, particularly in complex lighting conditions. A. M. Pujar and Poornima B. Kulkarni \cite{pujar2023automatic} developed an Automatic Number Plate Recognition (ANPR) system using R-CNN for character segmentation and OCR for recognition. Their system performs well under adverse weather, achieving high accuracy in segmenting and recognizing characters, making it suitable for traffic monitoring and toll collection. Vaibhav Pandilwar and Navjot Kaur \cite{pandilwar2023moving} integrated vehicle detection, speed estimation, and license plate recognition using YOLO and OpenCV. Their system achieved 98.3\% accuracy in license plate detection and an F1-score of 0.91\%, demonstrating effectiveness in diverse conditions, including partial occlusions and low lighting. Omar Gheni Abdulateef et al. \cite{abdulateef2022vehicle} presented a Deep Convolutional Neural Network (DCNN)-based model for vehicle license plate detection and localisation. The system employs anchor-free techniques to predict bounding quadrilaterals rather than traditional rectangles, improving the accuracy of license plate localisation. Through transfer learning, their model achieved a 98.8\% accuracy rate in vehicle classification and license plate detection on the Kaggle dataset. This system is particularly useful in applications such as traffic monitoring, toll collection, and vehicle classification in real-world scenarios.

\subsection{Speed Estimation}\label{chap:speedreview}
Relational vehicle speed data is crucial in traffic management and safety enforcement. Researchers have explored various approaches, each enhancing existing systems. Singh et al. \cite{singh2024enhancing} developed a hybrid deep learning model integrating CNN and Long Short-Term Memory (LSTM) networks for speed estimation. Originally applied to wind speed forecasting, this model adapted to vehicle speed estimation, improving prediction accuracy by combining CNN's spatial data capture with LSTM's temporal pattern proficiency. Eli Gilishish et al. \cite{gildish2024vibration} proposed a method for estimating rotational speed in Industrial Internet of Things (IIoT) systems using wireless vibration sensors, analyzing vibration signals without additional machine kinematics information. This method showed high accuracy for rotational machines like wind turbines by analyzing peak frequencies in vibration spectra. Paulo de Araujo et al. \cite{de2023novel} introduced a method for vehicle positioning and speed estimation using invariant Kalman filters and a deep-learning-based radar system, replacing accelerometer readings with radar data to minimize error accumulation. This approach measures forward speed and corrects noisy data with a deep-learning denoiser, achieving robust speed estimation and positioning suitable for autonomous vehicle navigation. Rodríguez-Rangel et al. \cite{rodriguez2022analysis} explored real-time vehicle speed estimation based on YOLO vehicle detection and Kalman filtering. Their system captures vehicle speeds using a monocular camera from the side of the road and utilizes YOLOv3 for vehicle detection, combined with Kalman filters for tracking. The Linear Regression Model (LRM) achieved the best performance, with a Mean Absolute Error (MAE) of 1.694 km/h for the centre lane and 0.956 km/h for the last lane. This method demonstrates high accuracy and low computational cost, making it suitable for real-time speed estimation applications using minimal infrastructure. 

\subsection{Gaps in Literature and Paper Contributions}
The existing literature reveals critical gaps, particularly concerning developing countries. There is a notable lack of research focused on regions like Uganda, where road safety infrastructure is limited and traffic fatalities are high. Most current systems operate in isolation, needing more integration of vehicle detection, license plate recognition, and speed estimation into a unified framework. Additionally, there is insufficient exploration of real-time data utilization for traffic enforcement, which could enhance responsiveness and effectiveness. Current models often must address scalability and adaptability to diverse environmental conditions typical of low-resource settings. Lastly, the literature overlooks the importance of user engagement in traffic enforcement systems, especially regarding feedback mechanisms that could improve compliance.

This study significantly contributes to intelligent traffic surveillance by proposing a comprehensive system tailored for Uganda. It integrates vehicle detection, license plate recognition, and speed estimation into a unified framework. The research provides actionable insights for local authorities to reduce traffic fatalities by focusing on practical applications to enhance road safety in Uganda. It also lays the groundwork for future studies by demonstrating the effective deployment of advanced technologies in developing regions. Finally, the findings inform policymakers about the benefits of integrating technology into traffic management strategies, promoting safer road environments.

\section{System Description}
\label{sec3}

\begin{figure*}
    \centering
    \includegraphics[width=\linewidth]{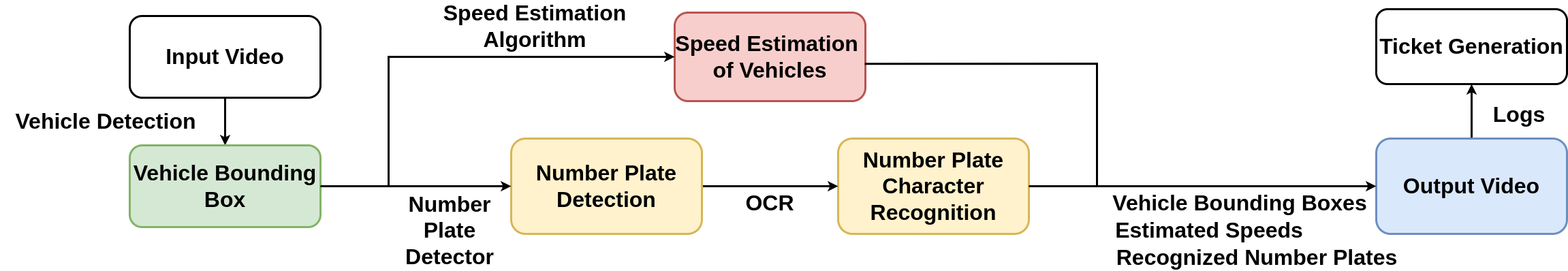}
    \caption{\textit{Overview of the flow system pipeline.}}
    \label{fig:flowchart}
\end{figure*}

Figure \ref{fig:flowchart} outlines the methodology used. The input video is processed to detect vehicles and annotate them with bounding boxes across all frames. License plates are then identified, and an algorithm matches each plate to its respective vehicle. The number plates are then sent to OCR for character recognition. Finally, vehicle speeds are estimated using transformation and distance-time calculations and compared with measurements from a speed gun. This integration facilitates logging, ticketing, and visual vehicle output. However, vehicle data were collected to implement this system due to the lack of publicly available datasets for speed estimation and plate detection in the Ugandan context.

\subsection{Data Collection and Preprocessing}
Data collection was conducted on various highways in Kampala, Uganda, in collaboration with local traffic officers, who guided the operation. Three devices were used for different data types: a speed gun (LT1 20/20 TruCAM II), a Canon EOS 5D Mark III camera, and a mobile phone with a 12 MP camera, as detailed in Table \ref{tab:device_specifications}. Each device served a specific purpose to achieve the project's objectives.

    \begin{figure*}
        \centering
        \includegraphics[width=0.8\textwidth]{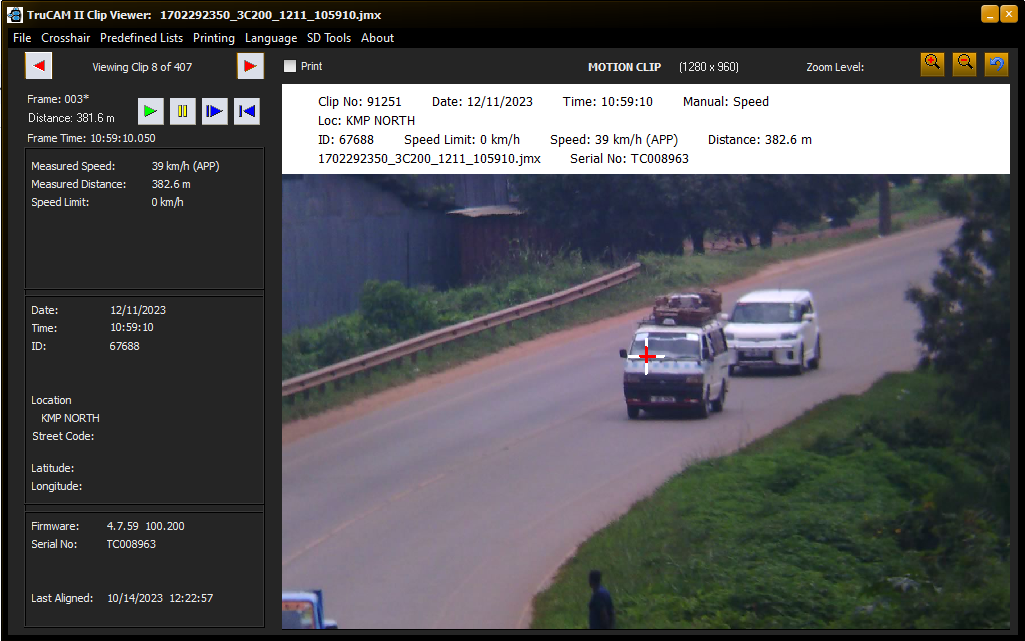}
        \caption[TruCAM viewer interface]{\textit{TruCam II Clip Viewer interface displaying captured vehicle snapshot details. The measured speed is 39 Km/h. }}
        \label{fig:trucam-viewer}
    \end{figure*}

\begin{itemize}
    \item \textbf{Speed Gun (LT1 20/20 TruCAM II):} This device was primarily used to capture snapshots of vehicle speeds, which served as true labels for evaluation. The speed gun recorded metadata for each snapshot, including serial number, timestamp, measured distance, measured speed, and the number of frames. The files could only be opened using the TruCAM II clip viewer, as shown in Figure \ref{fig:trucam-viewer}, provided by the Police. Each clip had a resolution of 1280 x 960 and contained fewer than eight frames (approx. 100 ms). This detailed metadata was crucial for correlating speed measurements with video footage captured by the Canon camera. For instance, if the speed gun recorded the speed of car A, its timestamp provided a central time point for extracting the corresponding video segment from the Canon camera's recordings. The total number was 407 clips.

    \item \textbf{Canon EOS 5D Mark III Camera:} The Canon camera recorded high-definition videos in .MOV format at a resolution of 1920 x 1080 pixels and a frame rate of 25 frames per second. Each video ranged from 30 seconds to 10 minutes in length, with an average of 2 minutes. The camera's primary role was to capture comprehensive footage of vehicles as they passed, enabling subsequent analysis of their speeds using the speed gun's snapshots. The recorded clips were trimmed based on timestamps from the speed gun to ensure that only relevant footage was used for speed estimation. A total of 51 videos were collected over 6 hours.

    \item \textbf{Mobile Phone:} The mobile phone captured images of vehicle number plates. Unlike the other devices, images from mobile phones were not used to estimate speed. However, they were essential for creating a dataset to train a license plate detection system. The phone's 12 MP camera provided sufficient resolution to capture clear images of number plates. Additionally, frames containing license plates were extracted from each video (captured by the Canon Camera) and combined with those from the mobile phone to form a dataset for training a license plate detector.
\end{itemize}

    \begin{table*}
        \centering
        \caption[Devices used during data collection and their corresponding specifications]{\textit{Devices used during data collection and their corresponding specifications}}
        \label{tab:device_specifications}
        \begin{tabular}{|l|l|}
        \hline
        \textbf{Device} & \textbf{Specifications} \\
        \hline
        Camera & Canon EOS 5D mark III, Res: 1920 x 1080, Filetype: .MOV, fps: 25 \\
        \hline
        Phone camera & iPhone SE, Camera: 12MP, Res: 3840 x 2160, Filetype: .MOV, fps: 60 \\
        \hline
        Speed gun & LT1 20/20 TruCAM II, Res: 1280 x 960, Filetype: .jmx \\
        \hline
        \end{tabular}
    \end{table*}

\subsection{Vehicle Detection and Tracking}
To detect vehicles, we utilized YOLOv8, which is designed to be fast, accurate, and easy to use, and is an excellent choice for vehicle detection. YOLO (You Only Look Once) partitions an image into a grid of cells and predicts bounding boxes and associated probabilities to detect vehicles. Given that road vehicles are among the classes in the COCO dataset, YOLOv8 was used out-of-the-box, with no fine-tuning, for our task \cite{sohan-2024, yaseen2024yolov8indepthexplorationinternal}. 
Given an image, YOLOv8 returns bounding boxes for every detected object, and, using a class definition, we extract the desired boxes, which are then passed to the tracker. For vehicle tracking, we initially used SORT \cite{wojke2017simpleonlinerealtimetracking} but later switched to ByteTrack \cite{zhang2022bytetrackmultiobjecttrackingassociating} because it associates each detection box with its corresponding object.

\subsection{License Plate Detection}
To accurately recognize a license plate, the plate's exact location must be detected. For this reason, detecting a license plate from an image or frame is the most crucial step of a system that involves plate detection. Robust, state-of-the-art algorithms are required for this step. To obtain the best results, we used two datasets: the first contains extracted images, and the second is a public dataset from a non-African context. The first dataset comprises approximately 475 images, cropped from frames extracted from the videos and carefully annotated. We apply preprocessing steps to increase the dataset size by exploiting variations in brightness, rotation, and exposure. The applied steps mimic the changes that could happen during system use. After applying these steps, we obtained a total of 1418 images. All these images were resized to 640 × 640. The public dataset is around 21173 images. The two datasets were combined to form the primary dataset. A YOLOv8 model was fine-tuned on the dataset for 50 epochs using an image size of 640 x 640.

\subsection{Optical Character Recognition (OCR)}
The other stage is character recognition. There are currently a lot of techniques applied to character and number recognition, most of which are specific to the English alphabet \cite{hamad2016detailed, shinde2012text}. Although such algorithms often work for some plates, they fail in most cases due to variable fonts, background noise, plate format variability, and other factors \cite{bazzo2020assessing}. Two approaches were employed: CNN-based and transformer-based. 

\begin{itemize}
    \item \textbf{CNN Approach:} Approach involved preprocessing images by converting them to greyscale and resizing them to 128 x 128 pixels. The dataset comprised approximately 2,430 license plate images. The CNN model included convolutional layers for feature extraction, specifically two Conv2D layers with 3x3 kernels followed by a pooling layer. These extracted features were then fed into an encoding layer consisting of dense layers. For the decoder, the encoded features are passed through two bidirectional LSTM layers. The final output layer utilized a dense layer with softmax activation. The model was trained using the Adam optimizer and a custom CTC (Connectionist Temporal Classification) loss function to manage variable-length character sequences, with a batch size of 32.

    \item \textbf{Transformer Approach:} The dataset used around 1217 images. The images were converted to RGB, and the labels were padded to a uniform length. The TrOCR processor \cite{li2022trocrtransformerbasedopticalcharacter} was defined to handle the loading, preprocessing, and encoding of images and labels. This approach uses a vision encoder to extract image features and a sequence-to-sequence decoder to generate text. The training involved passing the processed images and encoded labels to the model to learn the mapping between the image features and the corresponding number plate text.    
\end{itemize}

\begin{figure*}
    \centering
    \includegraphics[width=0.8\textwidth]{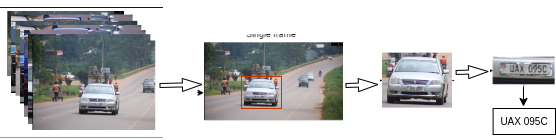}
    \caption{\textit{A frame from a video is used for vehicle detection; the output is then passed to the plate detection, whereby the characters are extracted using a recognition model.}}
    \label{fig:cropped_image_truth}
\end{figure*}

\subsection{Vehicle Speed Estimation}
\begin{figure}
    \centering
    \includegraphics[width=1\linewidth]{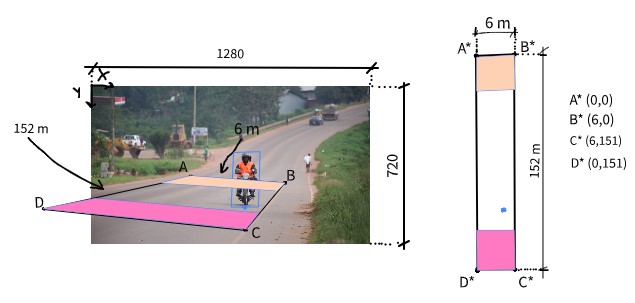}
    \caption{\textit{The source and target regions of the actual road scene. Points ABCD show the stretch of road as captured, while points A*B*C*D* show the transformed scene. This was crucial for speed estimation.}}
    \label{fig:perspective-transformation}
\end{figure}

As shown from figure \ref{fig:perspective-transformation}, first, source and target regions of interest (ROI) were defined in the video frames. The source ROI represents the actual road area, while the target ROI is a virtual region for mapping. Detections from the object detection model are filtered by confidence and class, then passed through ByteTrack to assign unique tracker IDs to each vehicle. The bottom-centre anchor point is transformed from the source ROI to the target ROI, and the resulting coordinates are stored in a double-ended queue for each tracker ID. To calculate speed, the algorithm computes the distance travelled by the vehicle within the target ROI by taking the absolute difference between starting and ending coordinates. Speed is derived by dividing distance by time, using perspective transformation to map real-world coordinates to a virtual space. Speeds are estimated across the region of interest, yielding multiple speed estimates for each tracked vehicle. To assign a single speed value, the maximum speed recorded during that stretch of road is used as the estimated speed. The algorithm can also determine the highest, the mode, or the lowest speed, depending on specific considerations.

\subsection{Ticket Generation}
The outputs from previous stages generate metadata used to identify each vehicle. This tracked metadata includes the detected number plate, estimated speeds within the region of interest, timestamp, and camera location. A report is generated based on this metadata to analyze individual vehicle information. To facilitate this, a dummy database was established to store vehicle-related user information. For instance, user A is associated with vehicle X and license plate ABC123A. The database includes the user's name, phone number, email, vehicle details, and license plate. The generated report is used to retrieve a vehicle corresponding to a specific license plate in the database, enabling user profile matching. Based on this information, further actions can be taken, including issuing an SMS ticket. To enable SMS functionality, Africa's Talking API was used to send tickets through text messages. To utilize SMS functionality, Africa's talking API was used to send tickets through text message \cite{africastalkingDocumentationAfricaaposs}.

\section{Results and discussions}\label{sec4}

\subsection{License plate detection evaluation}
License plate performance directly affects subsequent steps and the system's efficiency. The model was evaluated using the box loss, classification loss, detection loss, and mean average precision metrics. The mAP metrics are crucial for assessing object detection performance, as they quantify the model's ability to detect and localize objects accurately.

\begin{figure*}
    \centering
    \includegraphics[width=1\textwidth]{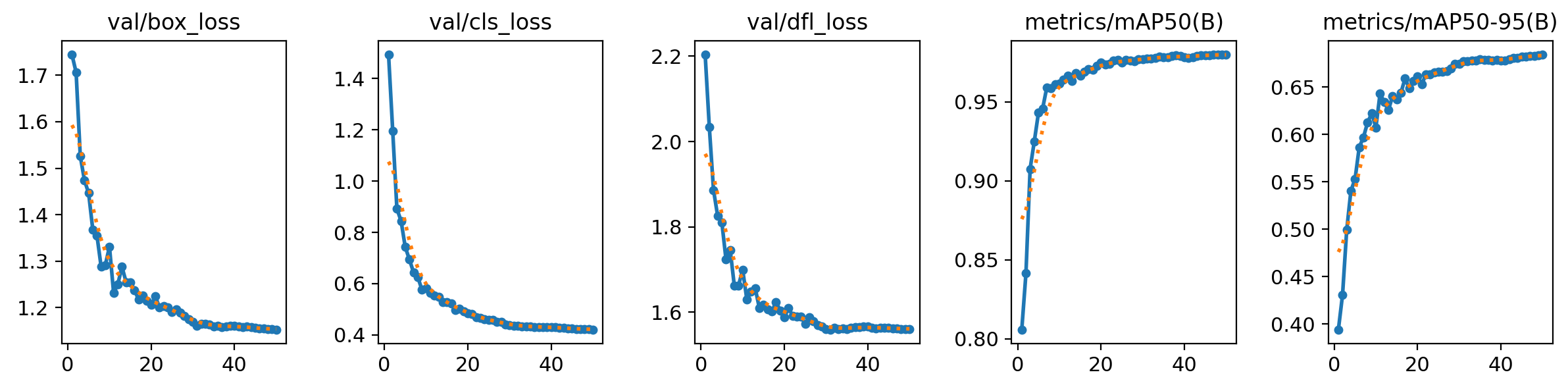}
    \caption{\textit{Training curves for the license plate detection model over 50 epochs, showcasing the progress of box loss, classification loss, detection filter loss, and mean Average Precision.}}
    \label{fig:trainedepochs_graphs}
\end{figure*}

\begin{table*}
\centering
\caption{Performance metrics for the license plate detection model trained for 50 epochs.}
\label{tab:various_model_metrics_plate}
\begin{tabular}{|c|c|c|c|c|c|}
    \hline
    \textbf{Metric} & val/box loss & val/cls loss & val/dfl loss & metrics/mAP50-95(B) & metrics/mAP50(B) \\
    \hline
    \textbf{Value} & 1.152 & 0.421 & 1.559 & 0.684 & 0.979 \\
    \hline
\end{tabular}
\end{table*}

As seen in table \ref{tab:various_model_metrics_plate}, the validation box loss is 1.152, the classification loss is 0.421, and the detection filter loss is 1.559. The low classification loss indicates that the model performs well at localizing plates. The mAP50 score of 0.979 suggests that the model performs well across varying levels, reliably identifying plates under clear conditions. Figure \ref{fig:trainedepochs_graphs} demonstrates how the loss values decreased over time, indicating the model was learning effectively.

\subsection{Character Recognition Evaluation}
\subsubsection{CNN Approach}
\begin{figure}
    \centering
    \includegraphics[width=1\linewidth]{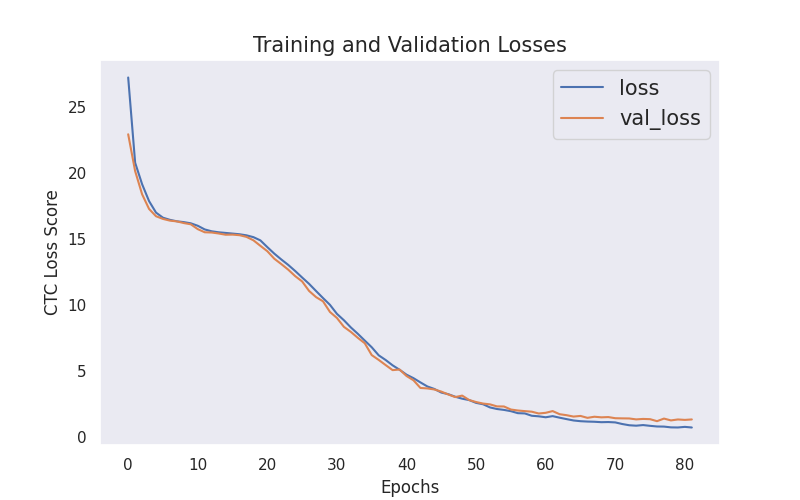}
    \caption{\textit{Training and validation loss curves. The dataset contained 2430 images.}}
    \label{fig:Training-and-Validation-Loss-Curve-2430-images}
\end{figure}

The model attained an overall mean character error rate (CER) of 0.038 on the test set. 
Figure \ref{fig:Training-and-Validation-Loss-Curve-2430-images} shows the loss curves of the CNN model with the achieved validation loss of 0.206.

\begin{table}
    \centering
    \setlength{\tabcolsep}{2mm}
    \renewcommand{\arraystretch}{1.2}
    \caption{Performance metrics of the CNN character recognition model.}
    \label{tab:cnn-char-rec-performance}
    \begin{tabular}{|c|c|c|c|c|}
        \hline
        \textbf{Parameters} & \textbf{Image count} & \textbf{Epochs} & \textbf{Eval loss} & \textbf{CER} \\ \hline
        \textbf{Value} & 2430 & 80 & 0.206 & 0.038 \\ \hline
    \end{tabular}
\end{table}

\begin{figure}
    \centering
    \includegraphics[width=1\linewidth]{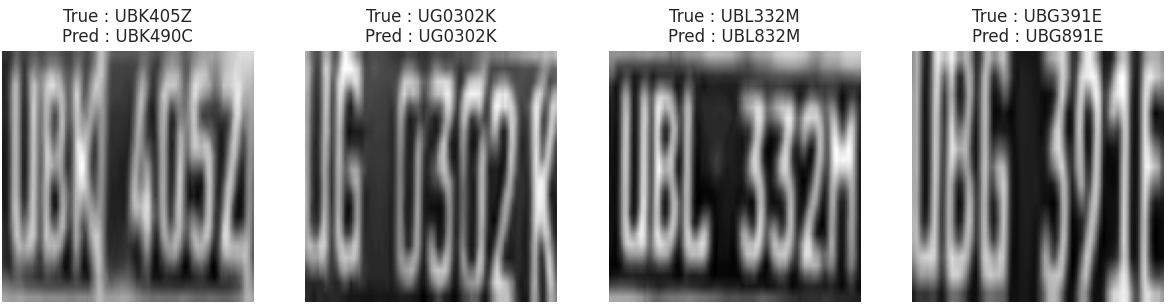}
    \caption{\textit{A sample of the predictions on the test data using the CNN model.}}
    \label{fig:test-images-on-cnn}
\end{figure}

The performance metrics for the CNN character recognition model are summarized in Table \ref{tab:cnn-char-rec-performance}. The model was trained on a dataset of 2430 images, achieving an eval loss of 0.206 and a character error rate of 0.038. The metrics suggest that the model effectively recognises characters, as indicated by the low CER.  Figure \ref{fig:test-images-on-cnn} illustrates some sample predictions. While the model performs well on the training data, it still struggles to generalize to unseen data, particularly for certain characters.

\subsubsection{Transformer Approach}

 The model achieved a CER of 0.0145 on the evaluation set. The model metrics, obtained on the 1217-image dataset, are reported in Table \ref{tab:train_test_metrics}.
 \begin{table*}
    \centering
    \caption{Performance metrics for the transformer model.}
    \label{tab:train_test_metrics}
    \begin{tabular}{|c|c|c|c|c|c|}
        \hline
        \textbf{Metric} & Epoch & Loss & Character Error Rate & Samples per second & Steps per second \\
        \hline
        \textbf{Train} & 10.0 & 0.153 & - & 2.72 & 0.34 \\
        \hline
        \textbf{Eval} & - & 0.144 & 0.018 & 3.294 & 0.414 \\
        \hline
    \end{tabular}
\end{table*}

\begin{figure}
    \centering
    \includegraphics[width=1\linewidth]{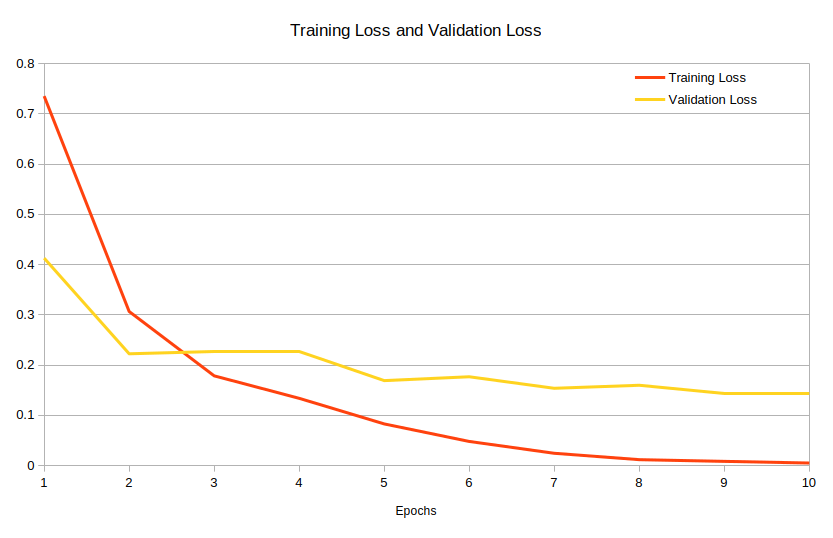}
    \caption{\textit{Training and validation loss curves on the image dataset.}}
    \label{fig:transformer-loss-curves}
\end{figure}

\begin{figure}
    \centering
    \includegraphics[width=1\linewidth]{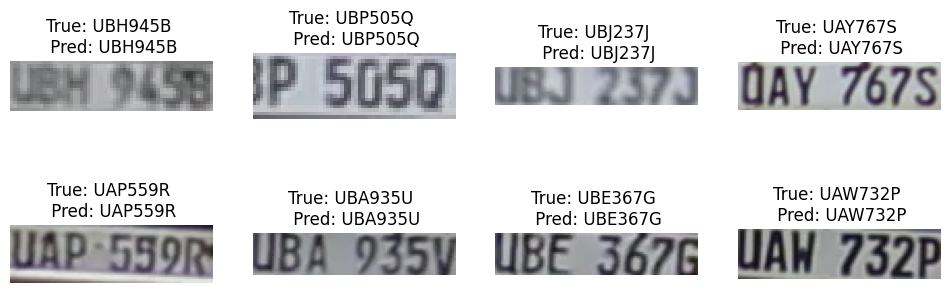}
    \caption{\textit{Predictions by the transformer model on a sample images.}}
    \label{fig:transformer-predictions}
\end{figure}

Figure \ref{fig:transformer-loss-curves} shows the training and validation loss curves for the transformer model, trained and evaluated on a dataset of 1217 images. The training loss curve shows a sharp downward trend, whereas the validation loss gradually increases. Additionally, 

Figure \ref{fig:transformer-predictions} presents sample predictions by the transformer showcasing its capability in character recognition.

While each approach shows promising results, it is worth noting some limitations. Transformers, though superior for character recognition, require either cloud-based inference or specialized hardware. The CNN was lightweight and easily deployable, though not accurate.  Table \ref{tab:model_comparison} presents the performance metrics of the best-performing CNN and transformer models, highlighting that the transformer, with only 10 training iterations, achieved a lower CER (0.0179) and evaluation loss (0.144) compared to the CNN, which required more iterations to reach a CER of 0.0385.

\begin{table*}
    \centering
    \caption{Comparison between the best-performing CNN and transformer models. Due to its attention mechanism, the transformer model requires significantly less training time.}
    \label{tab:model_comparison}
    \begin{tabular}{|c|c|c|c|c|}
        \hline
        \textbf{Architecture} & Iterations & Dataset images & Character Error Rate & Evaluation loss \\
        \hline
        \textbf{CNN} & 80 & 2430 & 0.0385 & 0.206 \\
        \hline
        \textbf{Transformer} & 10 & 1215 & 0.0179 & 0.144 \\
        \hline
    \end{tabular}
\end{table*}

\subsection{Speed Estimation}
The speed estimation results were validated against ground-truth data, indicating a reasonable level of performance. The algorithm successfully provided speed information for multiple vehicles simultaneously, although it exhibited an error margin of approximately ±10 km/h. This margin indicates that, although the system can effectively estimate vehicle speeds, there remains room for improvement in precision, particularly across varying environmental conditions.

\begin{figure}[t]
    \centering
    \begin{subfigure}{\linewidth}
        \centering
        \includegraphics[width=1\linewidth]{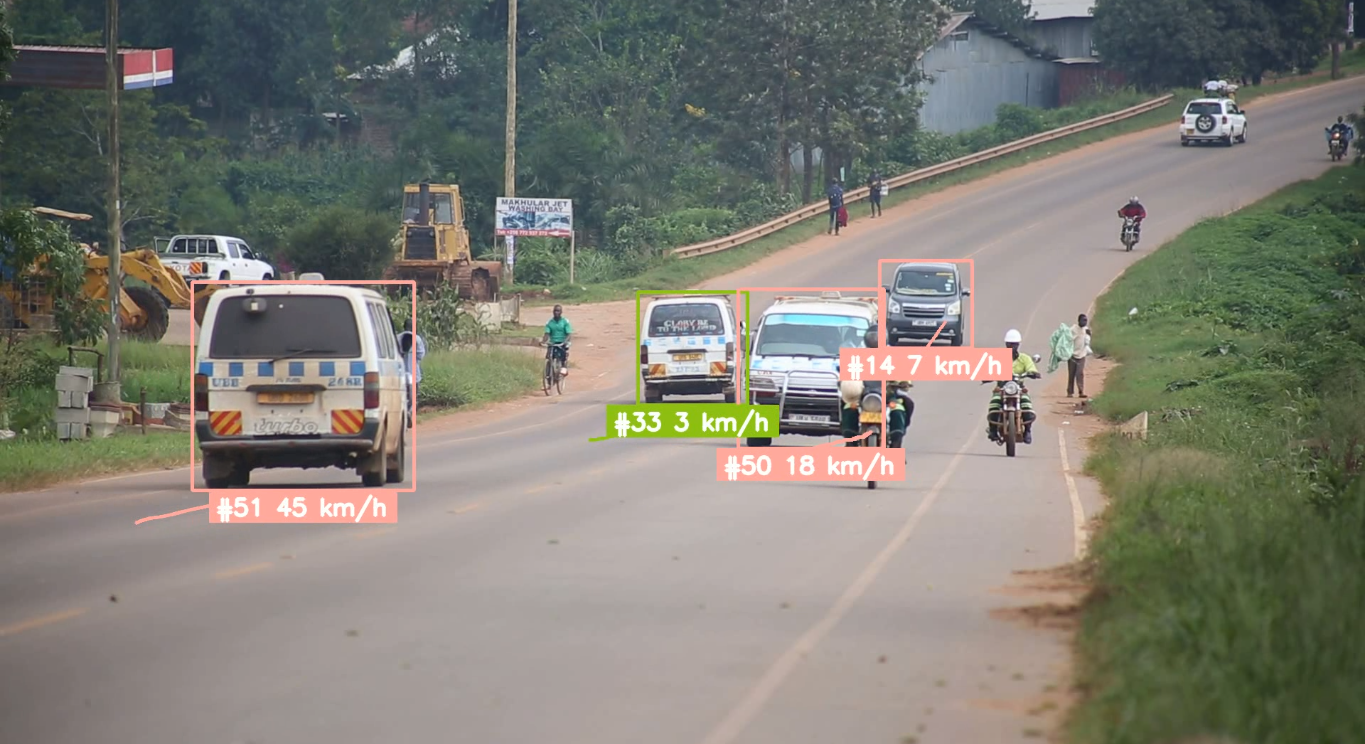}
        \caption{\textit{Sample picture of vehicles with estimated speeds}}
    \end{subfigure}

    \begin{subfigure}{\linewidth}
        \centering
        \includegraphics[width=1\linewidth]{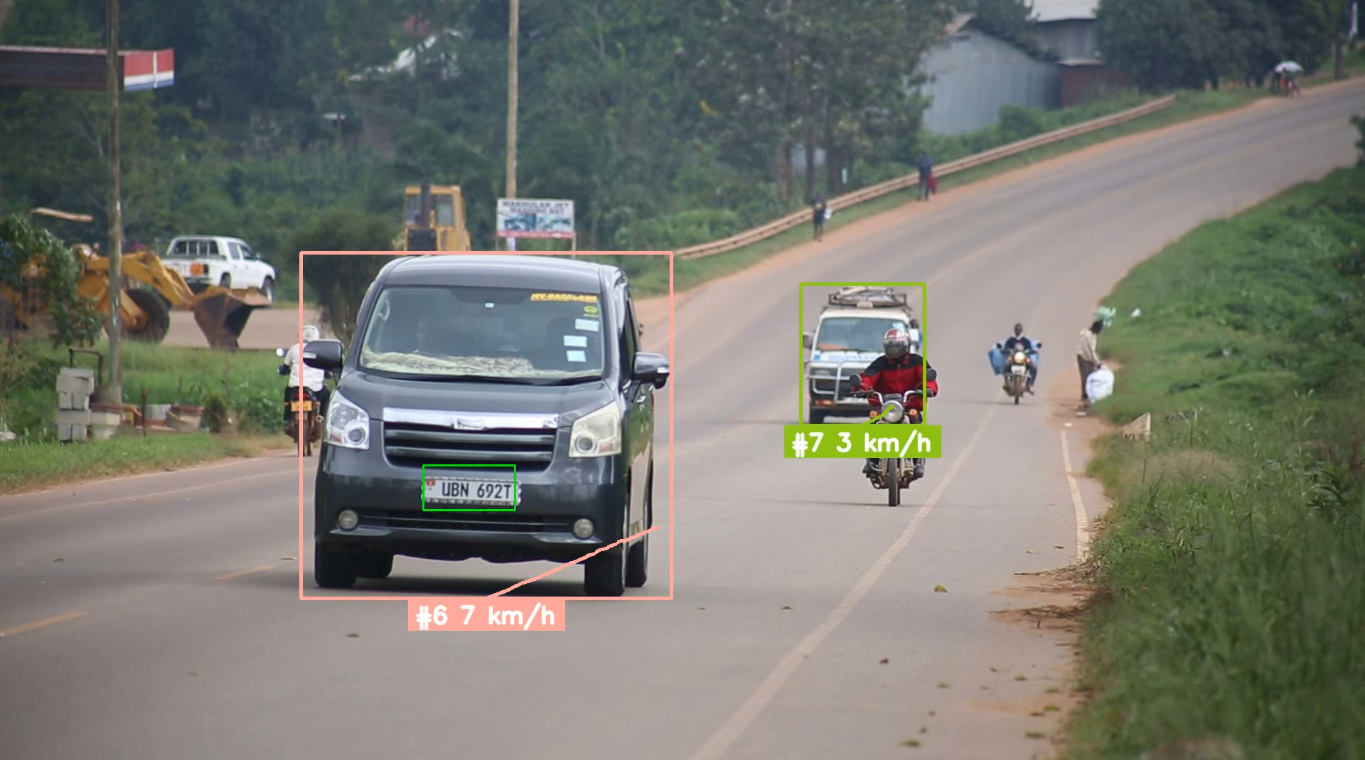}
        \caption{\textit{Fully processed sample of the frame from a video segment}}
    \end{subfigure}

    \caption{System Output Visualizations}
\end{figure}

\subsection{System Output}
The complete vehicle detection, tracking, and license plate recognition produces a comprehensive set of outputs for traffic surveillance. The comprehensive set of outputs from the different modules is integrated into a pipeline for data-driven decision-making and optimization. 
For each captured video, it produces 20-second output videos.

\begin{figure}[t]
    \centering

    \begin{subfigure}{\linewidth}
        \centering
        \includegraphics[width=1\linewidth]{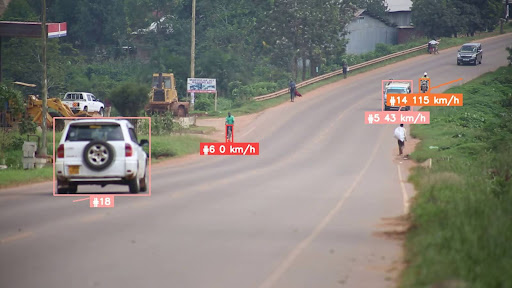}
        \caption{\textit{Vehicle with label \#5 annotated with the estimated speed of 43\,km/h.}}
        \label{fig:speed-est}
        \end{subfigure}

    \begin{subfigure}{\linewidth}
        \centering
        \includegraphics[width=1\linewidth]{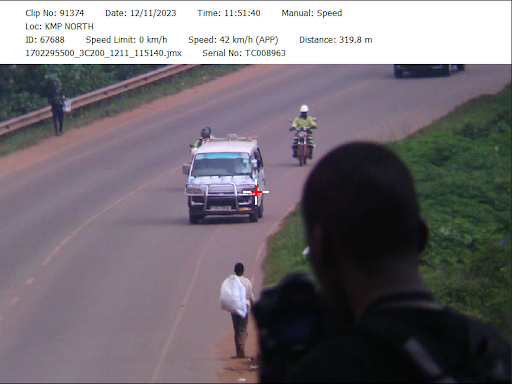}
        \caption{\textit{Same vehicle (\#5) captured by the speed gun with a speed of 42\,km/h.}}
        \label{fig:speed_estimate_comparison}
        \end{subfigure}

    \caption{Comparison between estimated vehicle speed and speed gun measurement.}
\end{figure}

Figures \ref{fig:speed-est} and \ref{fig:speed_estimate_comparison} show the exact vehicle captured by both the camera and the speed gun. This close agreement between the two measurements illustrates the algorithm's ability to estimate vehicle speeds to some extent, despite a significant error margin.

\subsubsection{Tickets Generated}
\begin{figure}
    \centering
    \includegraphics[width=1\linewidth]{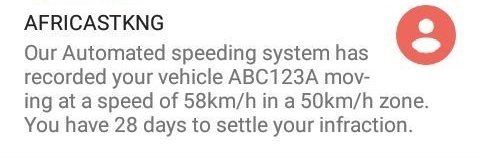}
    \caption{\textit{Sample ticket that is issued via SMS.}}
    \label{fig:issued_ticket3}
\end{figure}

Using the Africa's Talking API, the system was to send the ticket details seamlessly. This automation of SMS-based ticket delivery streamlines the overall enforcement process. Figure \ref{fig:issued_ticket3} shows a sample ticket issued.

\section{Conclusion}
The evaluation of various models and approaches in this study highlights the strengths of computer vision in developing systems that can benefit developing countries like Uganda. Our system achieved high mAP scores in license plate detection, particularly in ideal conditions, showcasing its effectiveness in real-time vehicle tracking. However, the occurrence of false positives and missed detections suggests a need for further refinement. The CNN model provided a competitive character recognition rate, though it struggled with certain characters, indicating the need for more comprehensive training data or potential model enhancements. However, the transformer model outperformed the CNN regarding accuracy and training efficiency but required specialized hardware for training, which might limit its practical use in resource-constrained environments. The speed estimation algorithm demonstrated reasonable accuracy but revealed a margin of error, underscoring the need for improved precision in speed calculations. Despite this, integrating these components into a unified system for traffic monitoring presents significant potential, especially for applications in Developing countries, where road safety remains a critical issue. Countries like Uganda's high traffic-related death rates, driven by speeding, highlight the importance of such systems in promoting safer roadways.
{
    \small
    \bibliographystyle{ieeenat_fullname}
    \bibliography{main}
}


\end{document}